\documentclass{article}

\usepackage{PRIMEarxiv}

\usepackage[utf8]{inputenc} 
\usepackage[T1]{fontenc}    
\usepackage{hyperref}       
\usepackage{url}            
\usepackage{booktabs}       
\usepackage{amsfonts}       
\usepackage{nicefrac}       
\usepackage{microtype}      
\usepackage{lipsum}
\usepackage{fancyhdr}       
\usepackage{graphicx}       
\usepackage{multirow}
\usepackage{caption}
\usepackage{subcaption}
\graphicspath{{media/}}     

\title{Remote Pathological Gait Classification System
}

\author{
  Pedro Albuquerque, Paulo Lobato Correia \\
  Instituto de Telecomunicações\\
  Instituto Superior Técnico, Universidade de Lisboa \\
  Portugal\\
  \texttt{pedro.flores.albuquerque@tecnico.ulisboa.pt, plc@lx.it.pt} \\
   \And
  João Machado, Luís Ducla Soares \\
  Instituto de Telecomunicações\\
  Instituto Universitário de Lisboa (ISCTE-IUL) \\
  Portugal\\
  \texttt{jpsmo11@iscte-iul.pt, lds@lx.it.pt} \\
   \And
  Tanmay Verlekar \\
  Dept. of CSIS and APPCAIR, BITS Pilani\\
  K K Birla Goa Campus\\
  Goa, India\\
  \texttt{tanmayv@goa.bits-pilani.ac.in} \\
  
}

\begin{document}
\maketitle

\begin{abstract}
Several pathologies can alter the way people walk, i.e. their gait. Gait analysis can therefore be used to detect impairments and help diagnose illnesses and assess patient recovery. Using vision-based systems, diagnoses could be done at home or in a clinic, with the needed computation being done remotely.
State-of-the-art vision-based gait analysis systems use deep learning, requiring large datasets for training. However, to our best knowledge, the biggest publicly available pathological gait dataset contains only 10 subjects, simulating 4 gait pathologies. This paper presents a new dataset called GAIT-IT, captured from 21 subjects simulating 4 gait pathologies, with 2 severity levels, besides normal gait, being considerably larger than publicly available gait pathology datasets, allowing to train a deep learning model for gait pathology classification. Moreover, it was recorded in a professional studio, making it possible to obtain nearly perfect silhouettes, free of segmentation errors. 
Recognizing the importance of remote healthcare, this paper proposes a prototype of a web application allowing to upload a walking person's video, possibly acquired using a smartphone camera, and execute a web service that classifies the person's gait as normal or across different pathologies. The web application has a user friendly interface and could be used by healthcare professionals or other end users. An automatic gait analysis system is also developed and integrated with the web application for pathology classification. Compared to state-of-the-art solutions, it achieves a drastic reduction in the number of model parameters, which means significantly lower memory requirements, as well as lower training and execution times. Classification accuracy is on par with the state-of-the-art.
\end{abstract}

\keywords{Assisted living \and Gait classification \and Pathology identification \and Remote diagnosis \and Web application}

\section{Introduction}
\label{sec:introduction}
Gait can be defined as the act of locomotion, involving periodic body movements, such as sequences of loading and unloading of the limbs \cite{kirtley2006clinical}. The study and analysis of gait in a medical context can contribute to the diagnosis and monitoring of pathologies that affect gait \cite{Boyd2005}. For this reason, the automatic analysis of pathological gait is gathering increased interest, with many approaches already available in the literature \cite{GaitAnalysisMethods, IdentifyingUsingAccelerometers}.

Of these approaches, vision-based solutions appear to be especially interesting, since image sequences can be captured with relatively simple setups, e.g. with a single 2D camera \cite{Verlekar2D}, which can easily be replicated in a clinical environment or even at home. Most of the processing required to analyze the observed gait can be done remotely, in a location where more computational power is available. A system based on this idea is considered in this paper to enable the remote classification of pathological gait.

Most state-of-the-art vision-based automatic gait pathology classification systems rely on deep learning \cite{GAIT-IST,NormalAndPathologicalLSTM, VerlekarUsingTransferLearning}. They involve the use of networks such as the VGG-19 \cite{VeryDeep} Convolutional Neural Network (CNN) model, pre-trained on the ImageNet dataset \cite{ImageNetDatabase}, and fine-tuned using (subsets of) pathological gait datasets. Fine-tuning requires relatively smaller datasets to adjust an existing model to perform better on a related problem. The quality of this adjustment and the expected results depend on the richness and suitability of the datasets used. However, most publicly available pathological gait datasets are captured from a limited number of healthy subjects simulating impaired gait conditions, as ethical and privacy concerns often prevent the sharing of data from real patients. 

This paper presents a new pathological gait dataset, GAIT-IT, containing more subjects, more simulated conditions and more image sequences compared to the state-of-the-art. It includes data from 21 subjects, simulating 4 gait related pathologies, with two severity levels, along with their normal gait. Sequences were acquired in a professional studio with a chroma-keying background, resulting in high a contrast between the foreground and the background. These characteristics are helpful for training a reliable deep learning model.

To validate the new dataset acquisition, a set of cross-validation tests, using GAIT-IT and the other publicly available datasets, is conducted. Results show that a model trained using GAIT-IT has improved generalization ability to make predictions over unknown data, confirming the usefulness of the proposed dataset for gait pathology classification studies.

The paper also presents a novel CNN architecture, which drastically reduces the number of trainable model parameters, compared to the state-of-the-art, thus having lower memory requirements and faster training and execution times.

A third contribution of this paper is a web  application for pathological gait classification. Given the increasing importance of remote healthcare, the proposed  web  application illustrates how remote diagnosis could be performed, with walking video sequences of an individual being remotely uploaded for analysis. The web service classifies the captured gait information across different pathologies, performing all computations on the server, and the results being returned to the individual in a user friendly manner. The web application's ease of use can make it effective for remote diagnosis and monitoring of gait related pathologies.

The rest of the paper is organized as follows. Section \ref{sec:Related Work} reviews gait representation models, pathological gait classification methods, and publicly available pathological gait datasets. Section \ref{sec:GAIT-IT} presents the proposed GAIT-IT dataset. Section \ref{sec:GaitClassificationWebApplication} describes the proposed web application and its gait pathology classification system, based on the developed low complexity CNN. Section \ref{sec:Performance Results} discusses the experimental results and Section \ref{sec:FinalRemarks} concludes the paper and presents directions for future work.

\section{Related Work}
\label{sec:Related Work}

Nowadays, a rich characterization of gait information can be obtained through the use of different types of sensors, including \cite{GaitAnalysisMethods}: i) floor sensors; ii) wearable sensors; and iii) vision sensors. 
Floor sensors, which can be used to detect ground reaction force measurements \cite{GroundForces} or the pressure exerted on each area under the foot \cite{GaitAnalysisMethods}, typically provide limited information for pathological gait classification and the equipment used is restricted to constrained spaces. On the other hand, wearable sensors are portable, allowing data acquisition of three-dimensional information related to walking patterns over long periods of time \cite{WearableSensors}, and can be used in many applications \cite{IdentifyingUsingAccelerometers}. However, their performance can be influenced by the sensor placement, which might also affect the subject's natural gait. 
Vision-based systems have the advantage of being unobtrusive and not requiring complicated subject cooperation. Currently, in this category, marker-based systems are considered as the gold standard approach for gait analysis. Such solutions \cite{MarkerBased} use special markers placed on key body parts to track them and obtain kinematic features from the observed motion. However, these often require specialized personnel to ensure setup and calibration processes and can be very time consuming. On the other hand, a markerless approach can be more suitable for application in less constrained environments \cite{2DMarkerless}, such as the integration of gait analysis in a clinical context. For these reasons, markerless vision-based systems are considered in this paper.

\subsection{Gait Representation}
\label{sec:Gait Representation}
In vision-based systems, the representations used for gait analysis typically follow a model-based or an appearance-based approach \cite{AReviewVisionBasedRecognition}.

In a \textbf{model-based approach}, gait representations are created by fitting a model to the input sequence of images or silhouettes, using prior knowledge of the human body (structural model) or its motion (motion model) \cite{VerlekarShadows}. An example includes two Kinect sensors with perpendicular viewing directions, acquiring RGB and depth to create a 3D model based on the movement of skeleton parts \cite{3DSkeletonJointsKinect}. This model combines static features (e.g., distances between joints), and dynamic features (e.g., speed, stride length or the body’s centre of mass movement).

An \textbf{appearance-based approach} represents gait without assuming prior knowledge of human motion. A sequence of binary silhouettes is typically obtained (e.g., by background subtraction), from which the desired gait representation is derived. 
A widely used representation is the Gait Energy Image (GEI) \cite{GEI}, obtained by averaging the cropped, normalized in size and horizontally aligned binary silhouettes of a gait cycle, according to Equation \ref{eq:GEI}:

\begin{equation} \label{eq:GEI}
    GEI(x,y) = \frac{1}{N}\sum_{i=1}^{N}B_{i}(x,y).
\end{equation}

\begin{figure}[!tb]
  \centering
  \includegraphics[width=0.5\textwidth]{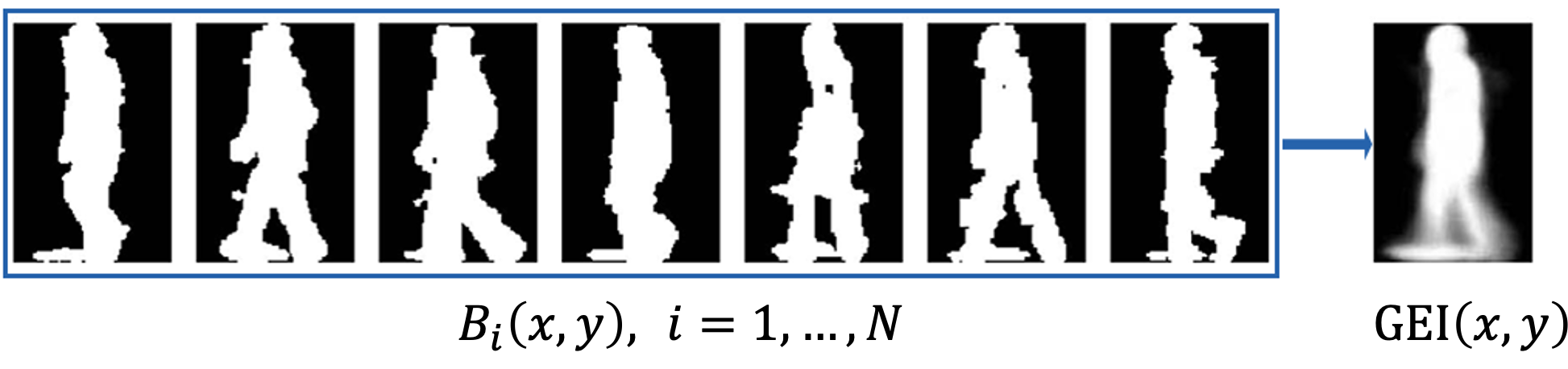}
  \caption[Example of binary silhouettes in a gait cycle and the corresponding GEI \cite{GEI}.]{Example of binary silhouettes in a gait cycle and the corresponding GEI \cite{GEI}.}
  \label{fig:FramesToGEI}
\end{figure}

\noindent$N$ represents the number of frames in one (or multiple) gait cycle(s) and $B_{i}(x,y)$ is a binary silhouette image, with $x$ and $y$ being pixel coordinates. The resulting GEI is a grey-level image implicitly representing, in a single image, the subject's shape and motion along the gait cycle. The GEI representation is robust against noise in individual frames, as illustrated in Figure \ref{fig:FramesToGEI}.

In this paper, results will be reported using the GEI for gait representation, as it is commonly used and provides a good compromise between representation power and computational efficiency.

A second representation considered for the presentation of results is the Skeleton Energy Image (SEI) \cite{GAIT-IST}, a hybrid between model- and appearance-based approaches. It starts by fitting a skeleton model to each image of the walking person, using OpenPose \cite{OpenPose}, as illustrated in Figure \ref{fig:back_sub_and_OpenPose}.b. With a skeleton image for each frame, the SEI can then be obtained with the same method used for GEI computation. The SEI was reported to achieve better pathological gait classification results than the GEI, as the SEI focuses on the dynamic movement characteristics and not on the physical constitution and clothing of a subject \cite{GAIT-IST}.


\subsection{Pathological Gait Classification}
\label{Pathological Gait Classification}
Classification of gait related pathologies from vision-based representations typically uses the visual gait representation directly, computes a set of biomechanical features or uses a combination of both. For instance, the work in \cite{NietoHidalgoClassification} describes two approaches, one using leg angles as features, and another one using the GEI.
A set of normalized gait features was proposed in \cite{ortells2018vision}, including the step length, stance and swing phases, or the amount and broadness of limb movements, to quantify gait impairments.

The last decade has witnessed the emergence of deep learning methods for feature extraction in image recognition and classification, including gait analysis systems. The solution presented in \cite{VerlekarUsingTransferLearning} adopts the GEI for gait representation and uses the VGG-19 model \cite{VeryDeep}, pre-trained on a subset of ImageNet \cite{ImageNetDatabase}, for feature extraction. Transfer learning was used to repurpose the model for pathological gait classification, with the last layers of the VGG-19 network being re-trained using GEIs computed from the INIT dataset \cite{ortells2018vision}. Linear Discriminant Analysis (LDA) was used for classification and the system's performance was tested using two other pathological gait datasets: DAI \cite{nieto2015vision} and DAI2 \cite{NietoHidalgoClassification}.
Another deep learning approach, also based on the VGG-19 model, was adopted in \cite{GAIT-IST} for pathological gait classification, using both GEI and SEI gait representations. In this case, the pre-trained model was fine-tuned  with data from the GAIT-IST dataset \cite{GAIT-IST}. 

Other deep learning approaches include the use of Recurrent Neural Networks (RNNs) that are able to learn correlations between inputs in a time series, such as the application of a bidirectional Long-Short Term Memory (LSTM) \cite{LSTM} network for pathological gait classification based on sequences of lower limb flexion angles \cite{NormalAndPathologicalLSTM}.

Given the good performance reported in the literature, this paper considers a deep learning solution based on the VGG-19 model as benchmarking for comparison against the new CNN model being proposed for gait analysis.

\subsection{Pathological Gait Datasets}
There are two types of gait datasets available, created either for gait recognition, or for pathological gait analysis. 
In \textbf{gait recognition datasets} subjects are required to walk normally, possibly including some covariates such as different speeds, different types of shoes, different clothing or carrying different items. Currently, a significant number of gait recognition datasets are publicly available. 
The purpose of \textbf{pathological gait datasets} is to include sequences of gait impaired due to some pathological condition, and there are much fewer publicly available datasets. Since sharing data from real patients raises ethical and data privacy issues, the publicly available pathological gait datasets were captured from healthy subjects simulating the characteristic gait impairments, after a learning and practice period.

Presently, there are four pathological gait datasets publicly available, as listed below. All the sequences in these datasets were captured from a canonical viewpoint and recorded in controlled environments.

The \textbf{DAI dataset} \footnote{\url{http://hdl.handle.net/10045/70567}} \cite{nieto2015vision}  contains binary silhouettes of 5 walking individuals. It has 15 normal gait sequences, and 15 sequences with random abnormal gait simulations, for a total of 30 gait sequences. The individuals are captured walking over a distance of 3 m using both the RGB camera of a Kinect sensor and a smartphone.

The \textbf{DAI2 dataset} \cite{NietoHidalgoClassification} also considers 5 walking individuals, but contains a total of 75 gait sequences. Each person simulates 4 pathologies (Parkinson's, diplegia, hemiplegia and neuropathy), as well as a normal walking gait. Each condition was recorded 3 times, while walking along a distance of 8 m.

The \textbf{INIT dataset} \footnote{\url{https://www.vision.uji.es/gaitDB/}} \cite{ortells2018vision} contains binary silhouettes of 10 individuals (9 males, 1 female), for a total of 80 sequences. Every subject is recorded 2 different times, at 30 fps, capturing multiple gait cycles and simulating seven different gait impairments (in addition to a normal gait sequence): i) right arm motionless; ii) half motion of the right arm; iii) left arm motionless; iv) half motion of the left arm; v) full body impairments; vi) half motion of the right leg; and vii) half motion of the left leg.

The \textbf{GAIT-IST dataset} \footnote{\url{http://www.img.lx.it.pt/GAIT-IST/}} \cite{GAIT-IST} considers 10 walking individuals, with a total of 360 gait sequences. The dataset includes the same 4 pathological gait types considered in DAI2, with 2 severity levels for each, 2 directions of walking, and 2 repetitions per participant, except for the normal gait. It is the largest pathological gait dataset currently available. Video sequences were captured using a smartphone camera with a resolution of $1280\times720$ pixels, mounted on a tripod at about 1.5 m above the ground and at a distance of about 4 m from the target.



\section{GAIT-IT Dataset}
\label{sec:GAIT-IT}

The goal of the proposed GAIT-IT dataset is to capture a larger gait pathology dataset, containing more variations. Having a more complete dataset, with higher quality images and a better contrast between the user and the background, allows us to obtain better models, which can generalize better to unknown data.

GAIT-IT was recorded in the professional studio of FCT$\vert$ FCCN (\emph{Fundação para a Ciência e a Tecnologia}) \footnote{\url{https://www.fccn.pt/en/collaboration/studio/}}, during two full days. The studio includes controlled artificial lighting and a green background, ideal for chroma-keying segmentation, allowing to compute high-quality binary silhouettes of walking subjects. Two professional 4K video cameras were used to capture synchronized gait sequences, one with a side view, at approximately 3 m from the target, and the other with a front/rear view, at about half a  meter from the walking start position. Both cameras stood on tripods at 1.75 m from the ground.

\subsection{Dataset Acquisition}
\label{sub:DatasetDescription}

The new GAIT-IT dataset includes sequences of normal gait and the same 4 pathological gait types present in the DAI2 and GAIT-IST datasets: diplegic, hemiplegic, neuropathic and Parkinsonian. For each pathology 2 levels of severity were considered, similarly to GAIT-IST, and the subjects were asked to provide 4 gait sequences per severity level and for their normal gait. This corresponds to a subject walking twice from left to right and from right to left, when imaged from the side view. The acquisition took place on two different days with the participation of 21 volunteers (19 males and 2 females) in the age range of 20-56 years old. Considering that sequences from 2 participants were acquired on both days, the GAIT-IT dataset includes a total of 828 gait sequences. Having sequences from 2 subjects acquired on different days, allows studying intra-subject variations, for instance due to wearing different clothes and shoes. 

Subjects were instructed on how to simulate the various gait types and severity levels, as summarized below \cite{StanfordSchoolofMedicine}. 

The \textbf{diplegic} pathology affects both sides of the body, with a forward leaning posture, and walking involves dragging both feet in a circular motion. For the second severity level the overall bending is accentuated, as well as leg and arm movements. 

The \textbf{hemiplegic} pathology affects only one side of the body (the right side was chosen). The leg is dragged in a circular motion, with a broader reach for the second severity, while the right arm remains still and held close to the waist, or flexed against the chest in the second severity level. 

The \textbf{neuropathic} pathology leads to foot drop and patients tend to lift their knees higher than normal to avoid dragging their toes on the floor. In the second severity level, the lift of the leg and the forward swing are exaggerated. 

The \textbf{Parkinsonian} pathology is characterized by a stooped posture, with the arms held close to the chest and the lower limbs flexed and rigid. Subjects were asked to attempt simulating general and erratic body shaking while taking small and relatively fast steps. The second severity level involved an overall exaggeration of these symptoms.

\subsection{Gait Representations Available in GAIT-IT}
\label{sub:Processing}

The GAIT-IT dataset provides various gait representations useful for gait pathology analysis: i) sequences of \textbf{binary silhouettes}; i) sequences of \textbf{skeletons}; iii) \textbf{GEIs}; and iv) \textbf{SEIs}. A GEI and SEI are available for each gait cycle, as well as for the complete set of gait cycles available per sequence. 

The spatial dimension of the produced gait representations is $224\times224$. However, the computation of gait representations is done with the full resolution of the captured gait sequences, to preserve information. Cropping removes the background around the subject's bounding box and then the width of the cropped image is padded to match its height, while maintaining the centroid  position. Finally, the square image is resized to $224\times224$ pixels, while maintaining the aspect ratio. All representations consider a 10 fps framerate. 

The main steps for obtaining the above gait representations are briefly described in the following.

\begin{figure}[tb]%
    \centering
    \subfloat{
    {\includegraphics[width=0.3\linewidth] {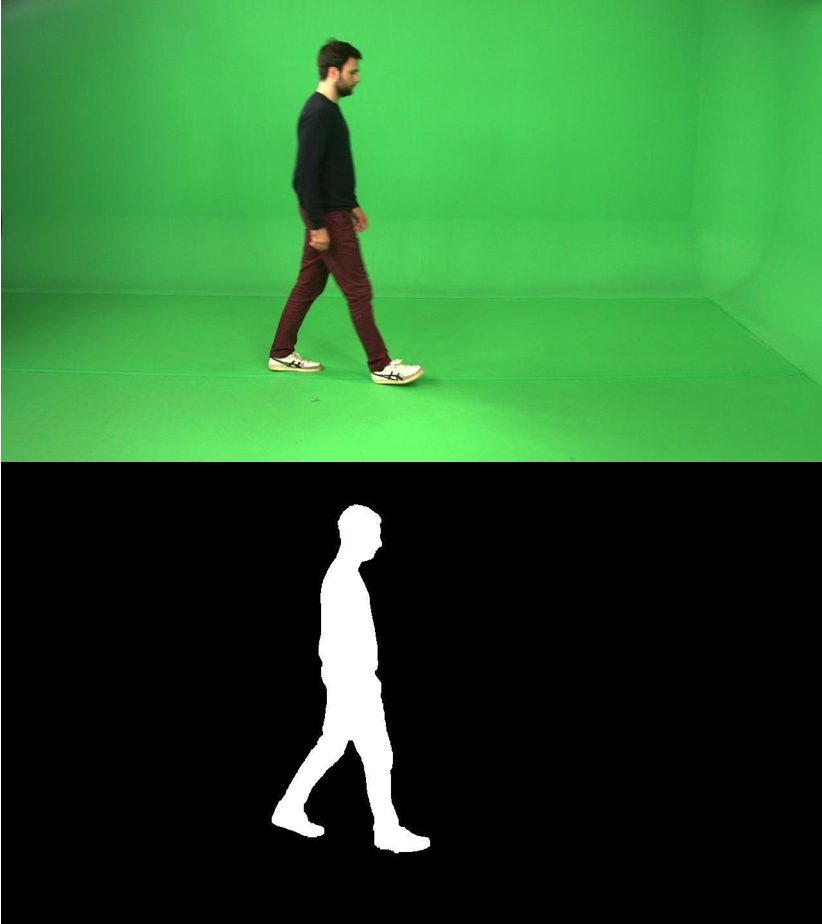}}
    }
    \quad
    \subfloat{\includegraphics[width=0.194\linewidth]{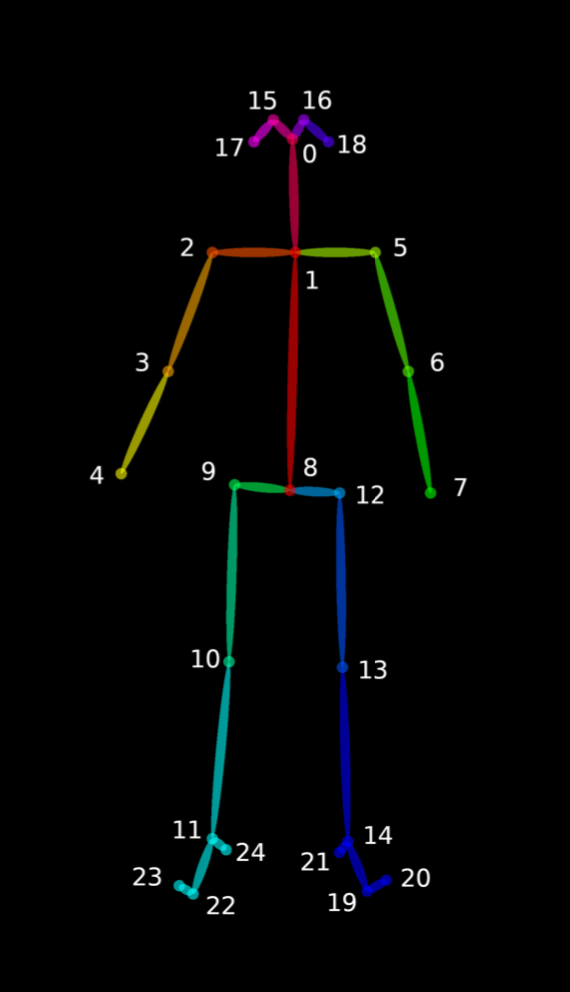}}
  \caption [(a) Side view silhouette obtained by background subtraction. (b) Pose output format of detected body parts using OpenPose \cite{OpenPose}.]{(a) Side view silhouette obtained by background subtraction. (b) Pose output format of detected body parts using OpenPose \cite{OpenPose}.}
    \label{fig:back_sub_and_OpenPose}
\end{figure}

The extraction of binary silhouettes relies on chroma-keying segmentation. A frame containing only the background is represented in the HSV colour space and the histograms of the hue (H), saturation (S) and value (V) components were computed. Then, all pixels in gait sequences with HSV values outside the background range are classified as belonging to the walking person's binary silhouette, and a morphological filtering operation is applied to remove small and isolated noise blobs. A sample result is included in the lower portion of Figure \ref{fig:back_sub_and_OpenPose}.a.

Skeleton computation relies on locating key anatomical parts in the gait images, using the  OpenPose software \cite{OpenPose}. OpenPose is able to automatically detect a total of 135 body, hand, facial and foot keypoints in each frame of a video, operating in real-time, using a multi-stage CNN. In this case, it was used to obtain the 2D coordinates of 25 keypoints corresponding to the full body, as illustrated in Figure \ref{fig:back_sub_and_OpenPose}.b \footnote{\url{https://github.com/CMU-Perceptual-Computing-Lab/openpose/blob/master/doc/output.md}}.

The computation of GEIs and SEIs follows the description provided in Section \ref{sec:Gait Representation}. The frames corresponding to a subject entering or leaving the camera's field of view were discarded, as well as the silhouettes and skeletons that did not correspond to a frame included in a complete gait cycle. 

\begin{figure}%
    \centering
    \subfloat{{\includegraphics[width=0.35\linewidth]{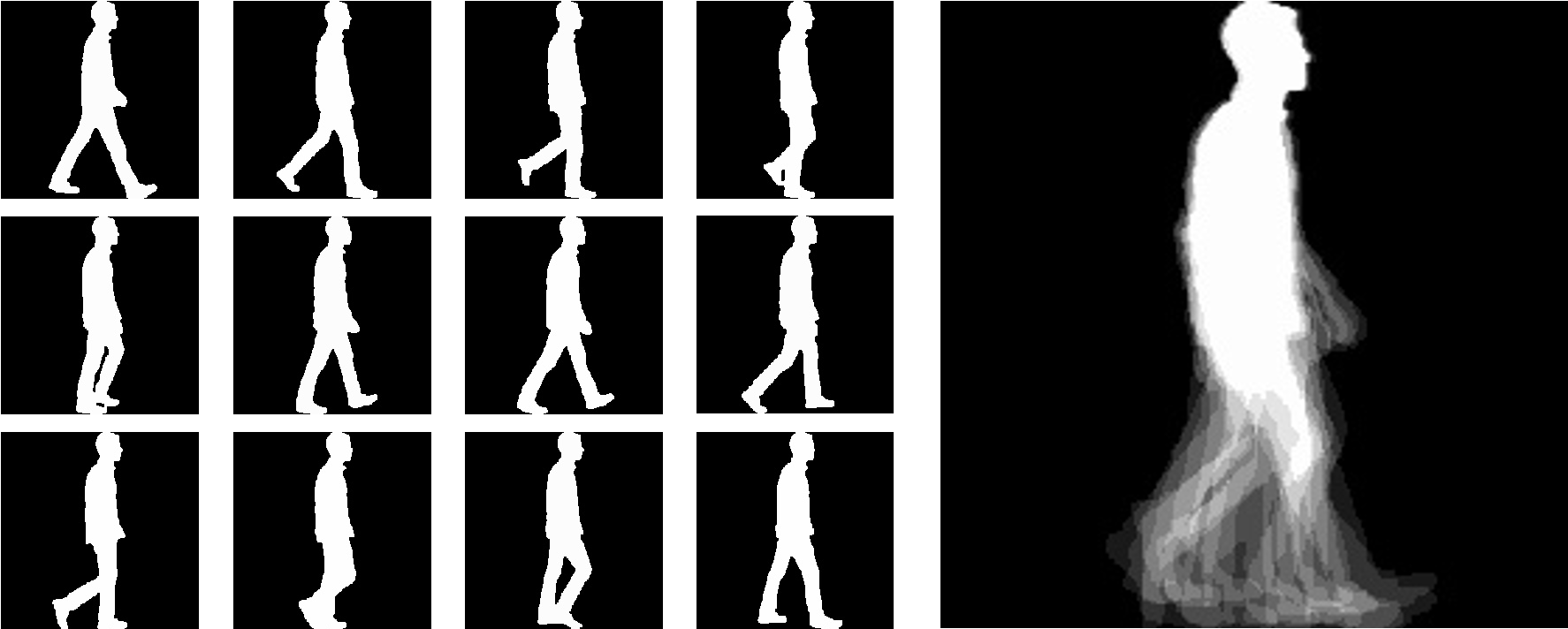}}}%
    \quad
    \subfloat{{\includegraphics[width=0.35\linewidth]{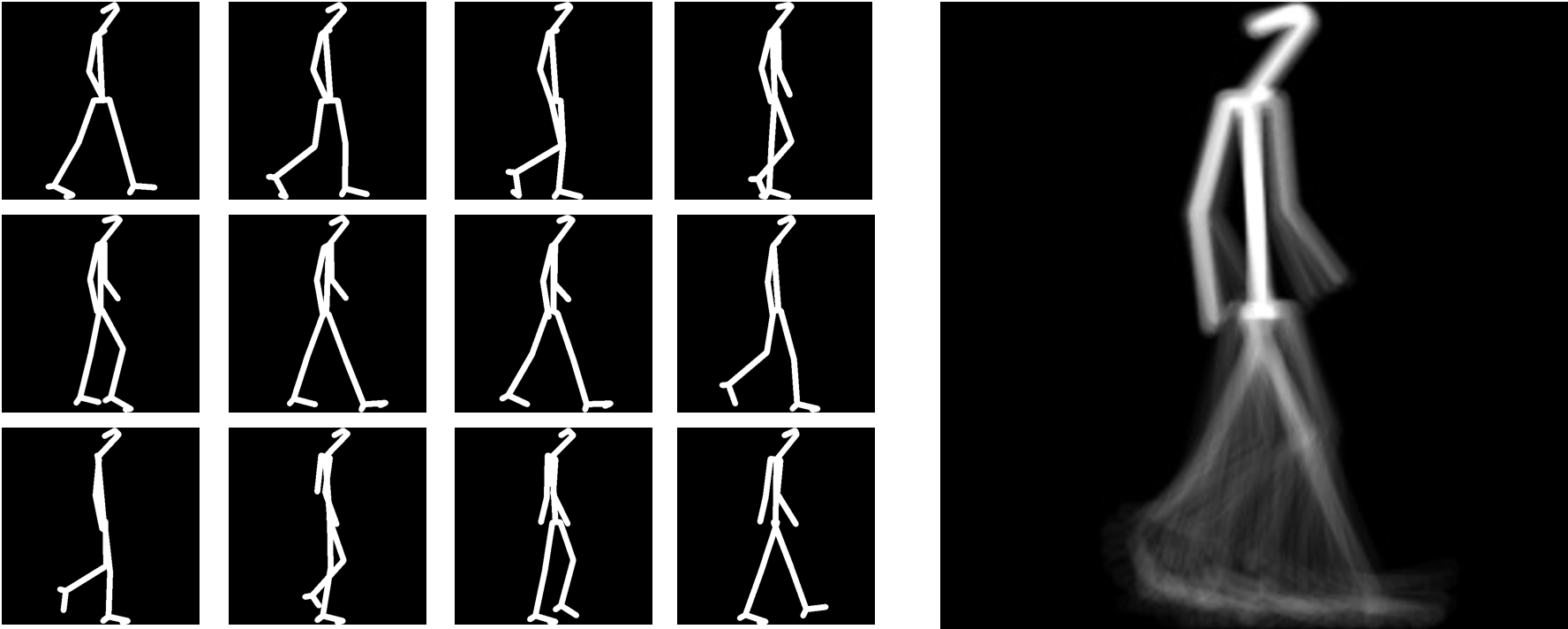}}}%
    \caption[Example of: (a) binary silhouettes in a gait cycle and the corresponding GEI; (b) binary skeletons in a gait cycle and the corresponding SEI.]{Example of: (a) binary silhouettes in a gait cycle and the corresponding GEI; (b) binary skeletons in a gait cycle and the corresponding SEI.}
    \label{fig:GAIT-ITGEIs}
\end{figure}

An example of the gait representations included in GAIT-IT, for one gait cycle, is included in Figure \ref{fig:GAIT-ITGEIs}.

\section{Gait Classification Web Application}
\label{sec:GaitClassificationWebApplication}

This paper proposes the prototype of a system that allows remote gait evaluation. It can assist healthcare professionals to identify patients requiring immediate attention and further examination, as well as to monitor the evolution of an existing gait pathology, without the need of physical interaction with the patient. The usefulness of such a system is made more evident under the Covid-19 pandemic.

The proposed web-based remote gait pathology classification prototype, is composed of two main modules: 

\begin{itemize}
    \item 
    \textbf{Automatic Gait Classification System} - This module accepts as input a gait representation, such as a GEI or SEI, and automatically classifies it as either normal or impaired  with one of the pathologies considered in the available datasets used for training: Parkinsonian, hemiplegic, diplegic or neurophatic gait; \item 
    \textbf{Web Interface} - This module provides an interface for access over the Internet, allowing the user to upload a gait video sequence, or directly a GEI or SEI, running the automatic classification system, and displaying the classification results in way that can be easily interpreted by the end user.
\end{itemize}

To better suit users with different degrees of expertise in using the automatic gait classification system, the web application provides two different interface modes. This allows advanced users to better understand the characteristics of the input gait sequence that contributed to the classification decision.

\subsection{Automatic Gait Classification System}
\label{sec:model_description}

The state-of-the-art vision-based systems rely on transfer learning to use a CNN model, such as the VGG-19 pre-trained on ImageNet \cite{VerlekarUsingTransferLearning,GAIT-IST}. Transfer learning can be especially important when dealing with small amounts of training data. It allows models trained for a different and somewhat related task to be adjusted to perform the new task, thus transferring the previously acquired knowledge to solve a new problem. However, even for transfer learning, a larger dataset can improve the quality of results obtained. 

The proposed pathological gait dataset, GAIT-IT, provides a considerable increase in the amount of available training data. Thus, rather than just using fine-tuning, it is now possible to develop a new light weight CNN specifically for the current classification task.

\begin{figure}[]
  \centering
  \includegraphics[width=0.7\linewidth]{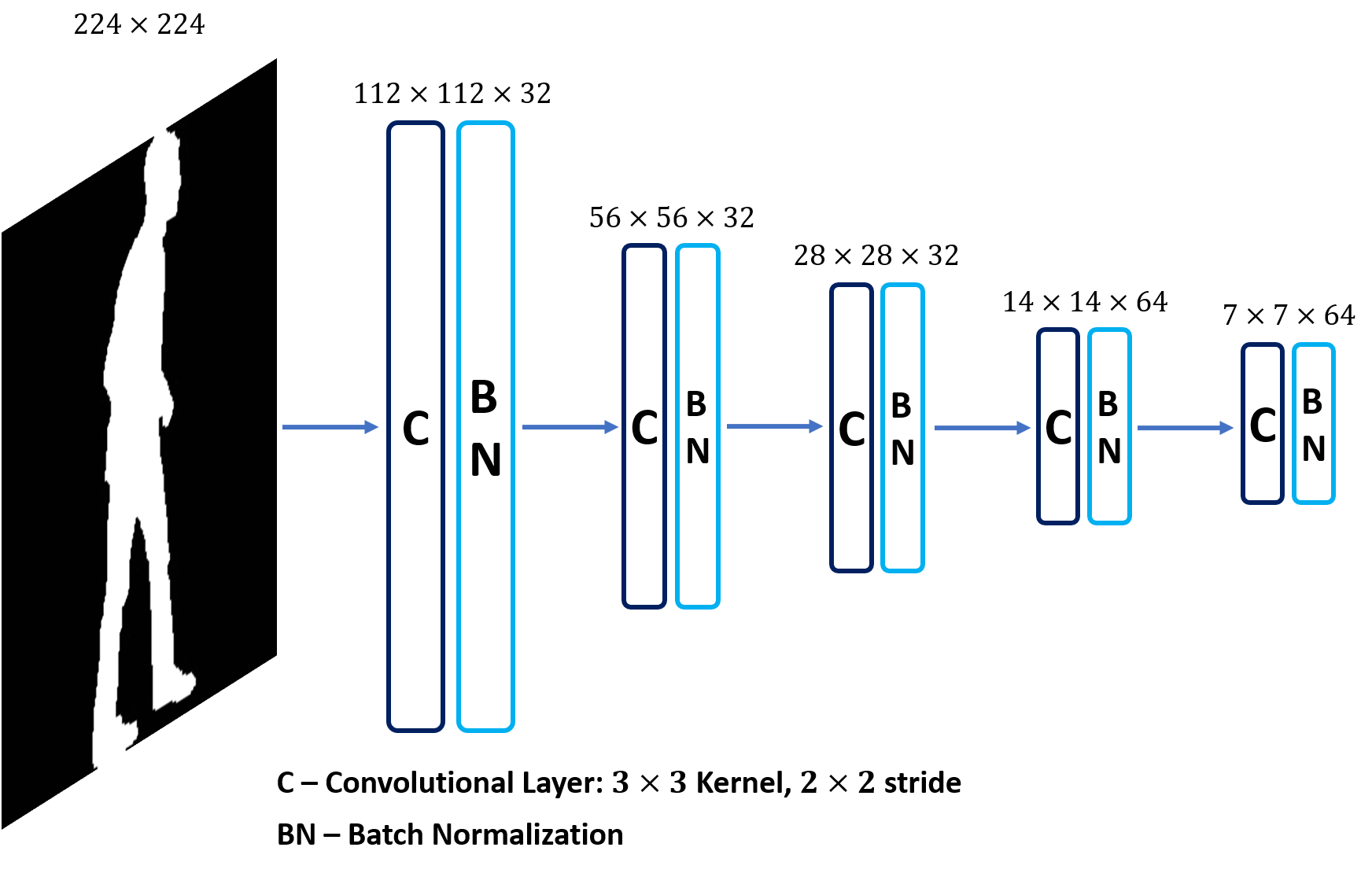}
  \caption[Illustration of the proposed CNN architecture for feature extraction in pathological gait classification.]{Illustration of the proposed CNN architecture for feature extraction in pathological gait classification.}
  \label{fig:CustomCNN}
\end{figure}

The architecture of the proposed CNN for feature extraction in pathological gait analysis is illustrated in Figure \ref{fig:CustomCNN}. Its main characteristics are:

\begin{itemize}
    \item \textbf{Network Depth}: The model includes 5 convolutional layers inspired by the networks designed for the Kaggle MNIST challenge \cite{kaggleMNIST} \cite{kaggleMNISTcompetitor}, which also consider binary input images.
    
    \item \textbf{Convolution Kernel}: In light of the work leading to the VGG CNN architectures, all the convolutional layers of the proposed CNN use small convolutional filters, with a receptive field of $3\times3$. A stride of $2\times2$ is considered.
    
    \item \textbf{Feature Maps}: The number of filters applied in each layer determines the number of feature maps at the layer's output. Starting with 32 filters at the image input, this number doubles for the last 2 layers.
    
    \item \textbf{Batch Normalization}: Each convolutional layer is followed by batch normalization, a layer that adjusts and scales its outputs to have a mean value close to 0 and a standard deviation close to 1. Bounding the values that pass between layers helps to stabilize and speedup the training process.
\end{itemize}

To perform classification, the features computed by the proposed CNN are flattened and submitted to a dense fully connected neural network, with two fully connected layers with a dropout \cite{dropout} of 0.5 between them. The first layer has 512 units, and the second has 5 units, corresponding to the 5 considered classes (normal, neuropathic, hemiplegic, diplegic or Parkinsonian gait), with a softmax activation to output class probabilities. This classification network is trained using categorical cross entropy as the loss function and the Adam algorithm \cite{Adam}, with the Nesterov momentum variation \cite{Nesterov1983AMF} and a learning rate of 0.001, as the optimizer.
\par

\subsection{Web Interface}
\label{sec:gait_web_application}

\begin{figure}[]%
    \centering
    \subfloat{{\includegraphics[width=0.45\linewidth]{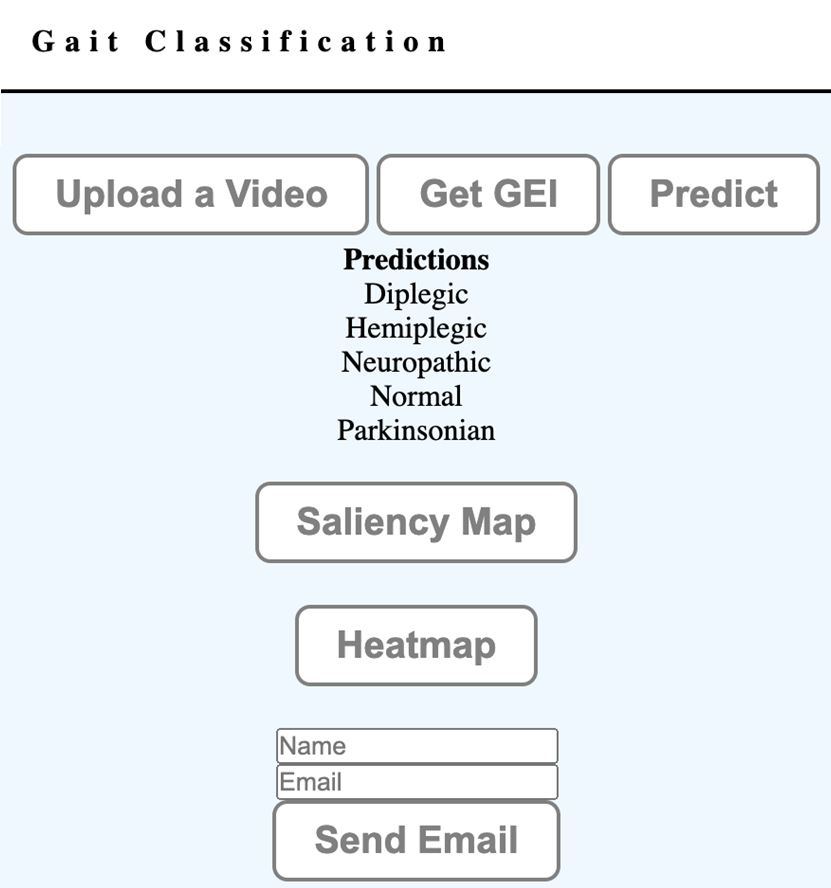}}}%
    \quad
    \subfloat{\includegraphics[width=0.45\linewidth]{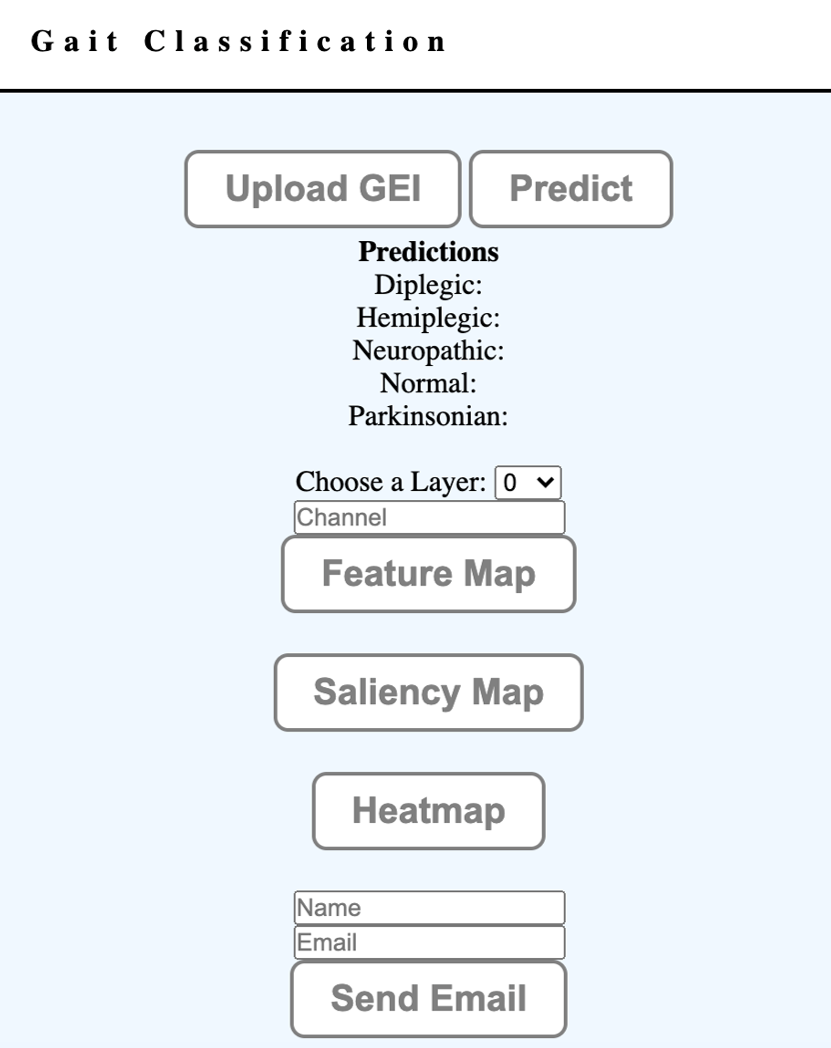}}%
    \caption[(a) Basic user interface mode. (b) Advanced user interface mode.]{(a) Basic user interface mode. (b) Advanced user interface mode.}
    \label{fig:web_interface}
\end{figure}

The web interface provides remote access, over the Internet, to the automatic gait pathology classification system described in Section \ref{sec:model_description}. The web application has two interface options:

\begin{itemize}
    \item \textbf{Basic} - The simpler interface could be used in a clinical environment, or even at home, where a simple setup for filming a walking person with any 2D camera, such as a smartphone camera, is available. The interface, illustrated in Figure \ref{fig:web_interface}.a, allows end-users to upload a video, and the web application computes a GEI representation of the observed gait and runs the automatic gait classification system. 
    Users can visualize the significance of the parts of the body that contributed to the classification process using saliency maps \cite{simonyan2013deep} and class activation maps (grad-CAM) \cite{selvaraju2017grad}. If the user so desires, the classification results can be sent to a specified e-mail address.
    This interface can be used to remotely obtain a preliminary diagnosis, or simply to help the healthcare staff to identify the regions that contribute most to the identified gait impairment.
    
    \item \textbf{Advanced} - The advanced interface, illustrated in Figure \ref{fig:web_interface}.b, could be used by researchers interested in analysing the operation of the deep learning classification solution, as they can visualise the feature maps generated by any of the CNN layers. This interface also allows to directly upload a previously computed GEI or SEI gait representation. The additional features of the advanced mode provide the users with an insight into the classification system operation.

\end{itemize}
 
The global Covid-19 pandemic has highlighted the importance of remote healthcare applications, to which the proposed web application prototype intends to contribute, providing the means to  obtain a preliminary diagnosis and help healthcare staff to identify patients in need of urgent attention. Running over the Internet, the proposed web application eases the access of remote populations and people with limited resources, only requiring an Internet connection and a simple 2D camera.\par
Since the proposed system deploys a web service, it means that also other web applications could access the gait pathology classification system, by issuing HTTP requests to the web service.
 


\begin{figure}[tb]%
    \centering
    \subfloat {{\includegraphics[width=0.2\linewidth]{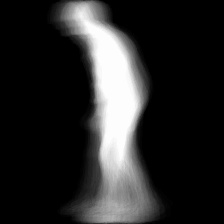}}}%
    \quad
    \subfloat {{\includegraphics[width=0.2\linewidth]{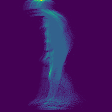}}}%
    \quad
    \subfloat {{\includegraphics[width=0.2\linewidth]{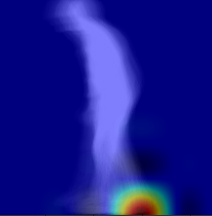}}}%
    \caption[(a) GEI example of hemiplegic gait; (b) Corresponding feature map of the first convolutional layer and twelfth channel; (c) Class activation map (grad-CAM).]{(a) GEI example of hemiplegic gait; (b) Corresponding feature map of the first convolutional layer and twelfth channel; (c) Class activation map (grad-CAM).}
    \label{fig:featureMapEx}
\end{figure}

\begin{figure}[tb]%
    \centering
    \subfloat {{\includegraphics[width=0.2\linewidth]{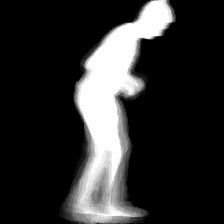}}}%
    \quad
    \subfloat {{\includegraphics[width=0.2\linewidth]{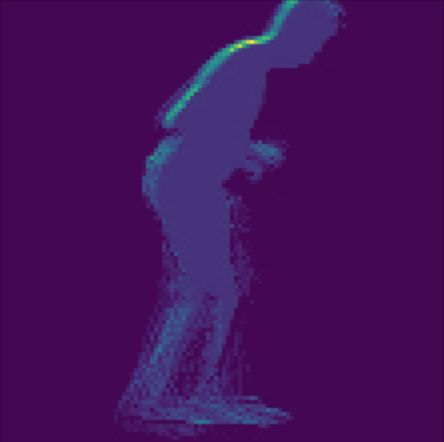}}}%
    \quad
    \subfloat {{\includegraphics[width=0.2\linewidth]{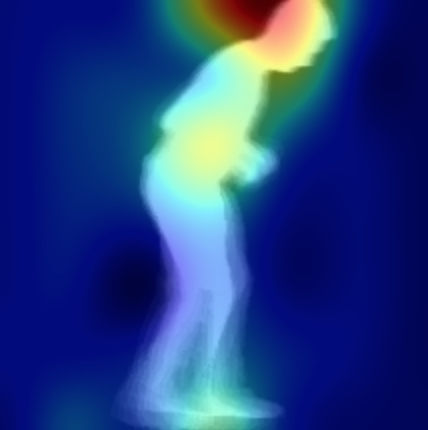}}}%
    \caption[(a) GEI example of Parkinsonian gait; (b) Corresponding feature map of the first convolutional layer and twelfth channel; (c) Class activation map (grad-CAM).]{(a) GEI example of Parkinsonian gait; (b) Corresponding feature map of the first convolutional layer and twelfth channel; (c) Class activation map (grad-CAM).}
    \label{fig:featureMapEx2}
\end{figure}

The advanced interface of the proposed web application gives users the possibility to access feature maps from different layers and channels, to visualize intermediate activation maps, as illustrated in Figures \ref{fig:featureMapEx}.b and \ref{fig:featureMapEx2}.b, displaying the feature map of the twelfth channel of the first convolutional layer, which appears to operate as an edge detector.

To further help understand the operation of the neural network, the web application also computes saliency \cite{simonyan2013deep} and gradient-weighted class activation maps (grad-CAMs) \cite{selvaraju2017grad}. These representations help users understand  which  input  image  areas  contribute  more  to the  CNN  classification  decision. As illustrated in Figure \ref{fig:featureMapEx}.c, the features highlighted when analysing an hemiplegic gait cycle are from the lower part of the body, namely related to the movement of the feet. On the other hand, for a Parkinsonian gait cycle, in which upper body movements are more affected, the proposed system paid more attention to the inclination of the head, the torso orientation and the position of the hands, as illustrated in Figure \ref{fig:featureMapEx2}.c. In the current implementation, these representations are obtained using the Keras Visualisation Toolkit \cite{raghakotkerasvis}. 


\section{Performance Results}
\label{sec:Performance Results}

The proposed gait classification system is evaluated using a 10-fold cross-validation protocol on the newly acquired GAIT-IT pathological gait dataset. To further emphasize the proposed system performance and generalization capability, a second test considers training on the GAIT-IT dataset and evaluation using GAIT-IST, thus performing a more challenging cross-database test. Detailed gait classification results, for each of the pathological gait types considered in GAIT-IT are reported in Section \ref{Performance Comparison On Different Gait Types}. \par

 It should be noted that, according to \cite{GAIT-IST} and \cite{VerlekarUsingTransferLearning}, among all the networks originally trained on Imagenet, VGG-19 performed the best, when fine-tuned to perform classification of gait related pathologies. Thus, to compare the proposed system with the state-of-the-art, the system presented in \cite{GAIT-IST} is fine-tuned using the proposed GAIT-IT pathological gait dataset. The fine-tuned system accepts a GEI or a SEI as input. As discussed in \cite{GAIT-IST}, when using GEIs as input, the last 3 convolutional blocks of the VGG-19 are fine-tuned, while for the SEI representation the best results are obtained by retraining all blocks except the first one.

\subsection{Cross-validation Results on GAIT-IT}
\label{Cross-validation Results on GAIT-IT}

The proposed CNN architecture and the state-of-the-art VGG-19 system considered for benchmarking purposes are evaluated following a 10-fold cross-validation protocol, on the GAIT-IT dataset. All GAIT-IT subjects are used, except the 2 subject repetitions. The test set for each fold is defined as $V_{k} = \{S_{i}, S_{{i+1}}, S_{i+2}\}$, where $i = 2\times k-1$, $k$ is the fold iteration and $S_i$ represents all sequences from one of the 21 available subjects, following the numbered labels used for each subject in the dataset. This arrangement ensures the use of all subjects in the test set at least once, while reducing the training bias.
Cross-validation results are presented in Table \ref{Cross-validation accuracies summary} as the average classification accuracy over all folds of the state-of-the-art system \cite{GAIT-IST}, as well as the classification model based on the proposed CNN. 
The low complexity CNN proposed here achieves a classification accuracy very close to the state-of-the-art classification accuracy, with 93.4\% and 92.6\%, when using GEI and SEI as inputs, respectively. 

However, the proposed CNN architecture, built specifically for the task at hand, has the major advantage of drastically reducing the number of trainable parameters. This allows to reduce the usage of static and dynamic memory to store and execute the system, as reported in Table \ref{Parameters} for the proposed CNN and the state-of-the-art benchmark VGG-19 system. Notice that the VGG-19 model is significantly larger than the one proposed in this paper, and converting the network models into the HDF5 \cite{HDF5} file format \cite{KerasAPI}, results in a model 83 times smaller than the state-of-the-art VGG-19 considered for benchmarking purposes.
This is an advantage of using the proposed system, which achieves a classification accuracy similar to VGG-19 using a shallower and low complexity CNN architecture.  
As a consequence of reducing the number of trainable parameters, the training process (i.e., fine-tuning of the network) is significantly faster. There is also reduction in the time needed to execute the classification system, which is of great importance when operating over the Internet. Table \ref{Duration} presents the time required for training and for executing the different networks. The entries represent the time taken to process one sample. These results suggest that the proposed CNN model is 15 times faster in training and 6 times faster in performing classification, when compared to the considered benchmark.

\begin{table}[t]
\caption[Cross-validation results using GAIT-IT dataset.]{Cross-validation results using GAIT-IT dataset.}
\label{Cross-validation accuracies summary}
\centering
\begin{tabular}{lcc} \hline
\textbf{Gait Classification System} & \textbf{Input Type} & \textbf{Accuracy} \\ \hline
VGG-19 fine-tuned \cite{GAIT-IST} & GEI & 94.0\% \\
Proposed CNN & GEI & 93.4\% \\
VGG-19 fine-tuned \cite{GAIT-IST} & SEI & 93.6\% \\
Proposed CNN & SEI & 92.6\% \\
\hline
\end{tabular}
\end{table}

\begin{table}[t]
\caption[Number of parameters and storage space required by the VGG-19 and the proposed CNN systems.]{Number of parameters and storage space required by the VGG-19 and the proposed CNN systems.}
\label{Parameters}
\centering
\begin{tabular}{lccc} \hline
\textbf{Gait Classification System} & \textbf{Input Type} & \textbf{Parameters} & \textbf{Size} \\ \hline
VGG-19 & GEI/SEI & 139,330,565 & 558.4 MB \\
Proposed CNN & GEI/SEI & 1,684,421 & 6.8 MB \\ \hline
\end{tabular}
\end{table}

\begin{table}[t]
\caption[Training and execution time (milliseconds) for VGG-19 and Proposed CNN.]{Training and execution time (milliseconds) for VGG-19 and Proposed CNN.}
\label{Duration}
\centering
\begin{tabular}{lccc} \hline
\textbf{Gait Classification System} & \textbf{Input Type} & \textbf{Train} & \textbf{Prediction} \\ \hline
VGG-19 & GEI/SEI & 15 & 6 \\
Proposed CNN & GEI/SEI & 1 & 1 \\ \hline
\end{tabular}
\end{table}

\subsection{Cross-dataset Tests on GAIT-IST}
\label{Cross-dataset Tests on GAIT-IST}
To further highlight the generalization ability of the proposed CNN model, cross-dataset tests were conducted using the GAIT-IST and GAIT-IT datasets. The proposed and the state-of-the-art systems were trained using all 23 subjects from GAIT-IT and the resulting model was tested using the 10 subjects from the GAIT-IST dataset.\par
Table \ref{Cross-dataset accuracies summary} reports the 10-fold cross-validation classification average results, obtained using both the GAIT-IT for training and GAIT-IST for testing. 
These results suggest that the proposed CNN system generalizes better than the state-of-the-art VGG-19 benchmarking system, when tested across different datasets. 

The proposed system achieves improved classification accuracy results over the state-of-the-art of 3.4\% and 1.3\%, when using GEIs and SEIs, respectively. This highlights that a deeper CNN architecture may be more prone to overfitting, and thus the shallower proposed CNN architecture is better suited for usage in a web application, which can receive gait video sequences captured in many different conditions, and using different types of video cameras.

\begin{table}[t]
\caption[Cross-dataset results using GAIT-IT for training and GAIT-IST for testing.]{Cross-dataset results using GAIT-IT for training and GAIT-IST for testing.}
\label{Cross-dataset accuracies summary}
\centering
\begin{tabular}{lcc} \hline
\textbf{Gait Classification System} & \textbf{Input Type} & \textbf{Accuracy} \\ \hline
VGG-19 & GEI & 86.4\% \\
Proposed CNN & GEI & 89.8\% \\
VGG-19 & SEI & 85.1\% \\
Proposed CNN & SEI & 86.4\% \\ \hline
\end{tabular}
\end{table}

\subsection{Performance Comparison on Different Gait Types}
\label{Performance Comparison On Different Gait Types}
The proposed system classifies gait across 5 different classes, which include 4 gait related pathologies and normal gait. Apart from obtaining state-of-the-art classification results, the proposed system aims to achieve a balanced performance across the different gait related pathologies considered. To confirm that this goal was achieved, Table \ref{Classification Gait Type} presents the obtained classification accuracies for each gait type on GAIT-IT. \par
To further analyze the system's performance with respect to each gait type, a confusion matrix is presented in Table \ref{Average Confusion Matrix}, including the average of the classification results obtained using both the GEI and SEI representations.  
The confusion matrix highlights common mislabeled predictions, providing valuable insight to analyze results, for instance highlighting that there are some  similar gait impairments that appear in different gait related pathologies. \par
From the presented results, normal gait appears as the easiest to classify, with a classification accuracy of 99\% using the proposed CNN model. Diplegic gait is the most difficult to classify, with a classification accuracy of 89\%, sometimes being incorrectly misclassified as hemiplegic or Parkinsonian. Hemiplegic gait performs slightly better with an average classification accuracy of 92\%. The distinct walking pattern of neuropathic gait allows the system to achieve an average classification accuracy of 97\%.  Parkinsonian gait achieved the next best classification accuracy of 95\%. These results suggest that most misclassifications are among diplegic and hemiplegic gait, as both pathologies are characterized by a stooped posture and relatively small step lengths. 

\begin{table}[t]
\caption[Classification accuracy across 5 different gait types.]{Classification accuracy across 5 different gait types.}
\label{Classification Gait Type}
\centering
\resizebox{\linewidth}{!}{%
\begin{tabular}{lcccccc}
\hline
\textbf{Gait Classification} & \multicolumn{1}{l}{\textbf{Input}} & \multicolumn{5}{c}{\textbf{Gait Type}} \\ 
\cline{3-7}
\multicolumn{1}{l}{\textbf{System}} & \textbf{Type} & \textbf{Diplegic} & \textbf{Hemiplegic} & \textbf{Neuropathic} & \textbf{Normal} & \textbf{Parkinsonian} \\ \hline
VGG-19 & GEI & 0.89 & 0.90 & 0.97 & 0.99 & 0.95 \\
Proposed CNN & GEI & 0.89 & 0.86 & 0.96 & 0.99 & 0.96 \\
VGG-19 & SEI & 0.86 & 0.91 & 0.98 & 0.99 & 0.95 \\
Proposed CNN & SEI & 0.87 & 0.88 & 0.97 & 0.98 & 0.93 \\ \hline
\multicolumn{2}{l}{Average} & 0.88 & 0.89 & 0.97 & 0.99 & 0.95 \\ \hline
\end{tabular}}
\end{table}

\begin{table}[t]
\caption[Confusion matrix detailing the classification results for each gait type, averaged between both systems in Table \ref{Classification Gait Type} using GEI and SEI inputs.]{Confusion matrix detailing the classification results for each gait type, averaged between both systems in Table \ref{Classification Gait Type} using GEI and SEI inputs.}
\label{Average Confusion Matrix}
\centering
\resizebox{\linewidth}{!}{%
\begin{tabular}{llccccc} \hline
 &  &  &  & \textbf{Predicted Class} &  &  \\ \cline{3-7}
& \multicolumn{1}{c}{\textbf{Class}} & \multicolumn{1}{c}{Diplegic} & \multicolumn{1}{c}{Hemiplegic} & \multicolumn{1}{c}{Neuropathic} & \multicolumn{1}{c}{Normal} & \multicolumn{1}{c}{Parkinson} \\ 
\hline
\multirow{5}{*}{\rotatebox[origin=c]{90}{\textbf{True Class}}} & \multicolumn{1}{l}{Diplegic} & \textbf{0.87} & 0.07 & 0 & 0 & 0.05  \\
& \multicolumn{1}{l}{Hemiplegic} & 0.09 & \textbf{0.89} & 0.02 & 0 & 0 \\
& \multicolumn{1}{l}{Neuropathic} & 0 & 0.02 & \textbf{0.97} & 0.01 & 0 \\
& \multicolumn{1}{l}{Normal} & 0 & 0 & 0 & \textbf{0.99} & 0 \\
& \multicolumn{1}{l}{Parkinsonian} & 0.05 & 0 & 0 & 0 & \textbf{0.95} \\ \hline
\end{tabular}}
\end{table}

\section{Final Remarks}
\label{sec:FinalRemarks}
This paper proposes a remote healthcare system for obtaining an automatic gait pathology assessment, presenting the prototype of a web application that can be used over the Internet. The web application implements a web service that executes a deep learning gait pathology classification system, and reports results using a friendly graphical user interface.
A novel gait pathology classification system based on a shallow CNN architecture was proposed, which performs at the same level as the state-of-the-art classification systems available. However, the developed system has two clear advantages over the state-of-the-art: (i) the proposed architecture has 83 times less parameters than an architecture based on VGG-19, with the corresponding advantages in terms of memory requirements, as well as training and testing times; (ii) The shallower network is less prone to overfitting, as confirmed by the cross-database tests reported, which is a major benefit when operating over the Internet and accepting gait video sequences acquired in very different conditions.
To allow a more complete training and testing of the proposed classification system, a new and larger pathological gait dataset, GAIT-IT, was acquired. GAIT-IT contains 828 gait sequences, featuring 21 subjects simulating 4 different gait related pathologies, namely, diplegia, hemiplegia, neuropathy and Parkinson’s disease, besides normal gait. GAIT-IT makes available 4 types of gait representations, notably silhouettes, skeletons, GEIs and SEIs. GAIT-IT sequences are recorded using two synchronized cameras, capturing both the sagittal and frontal points of view.

Since this work focuses on the sagittal view, future work can consider integrating also frontal view analysis. The combination of orthogonal view points will result in more meaningful information, leading to an improved classification system. Furthermore, alternative system architectures allowing the simultaneous processing of multiple synchronously acquired sequences \cite{LSTMGateFusion} can be considered to substitute the proposed CNN classification system. The web application can be further improved to allow training the system with new pathologies and additional representations in the advanced mode.

\section*{Acknowledgments}
This work was partly funded by FCT/MCTES under the project UIDB/EEA/50008/2020.

\bibliographystyle{unsrt}  
\bibliography{references}

\end{document}